\documentclass{article}
\usepackage{arxiv}  
\usepackage{graphicx}

\def \GAA {R2-AD2}
\def \AAA {A\textsuperscript{3}}

\usepackage[table]{xcolor}
\usepackage{standalone}
\usepackage{graphicx}
\usepackage{subfig}
\usepackage{tikz}
\usepackage{pgfplots}
\pgfplotsset{compat=1.16}

\usepackage{amsmath,amssymb,amsfonts}
\usepackage{cleveref}
\usepackage{enumitem}

\usepackage{tabularx}
\usepackage{makecell}
\usepackage{booktabs}
\usepackage{rotating}
\usepackage{tablefootnote}

\usepackage{array}
\newcolumntype{H}{>{\setbox0=\hbox\bgroup}c<{\egroup}@{}}

\usepackage{algorithm2e}
\SetKwComment{Comment}{// }{}

\usepackage{abbr}
\usepackage{style}

\usepackage{pgfkeys,pgfmath}  
\usepackage{siunitx}
\usepackage{expl3}
\ExplSyntaxOn
\newcommand{\fpabs}[1]{\fp_eval:n{abs(#1)}}
\ExplSyntaxOff
\newcommand{\percincrease}[3][0]{%
    \pgfmathdivide{#3}{#2}%
    \pgfmathparse{\pgfmathresult-1}%
    \pgfmathmultiply{\pgfmathresult}{100}%
    \SI[round-mode=places,round-precision=#1]{\pgfmathresult}{\percent}
}%

\begin{document}
\title{\GAA{}: Detecting Anomalies by Analysing the Raw Gradient}
\author{
	Jan-Philipp Schulze \\
	Fraunhofer AISEC \\
	\texttt{jan-philipp.schulze\thanks{\texttt{@aisec.fraunhofer.de}}} \\
	\And
	Philip Sperl \\
	Fraunhofer AISEC \\
	\texttt{philip.sperl\footnotemark[\value{footnote}]} \\
	\And
	Ana Răduțoiu \\
	Fraunhofer AISEC \\
	\texttt{ana.radutoiu\footnotemark[\value{footnote}]} \\
	\And
	Carla Sagebiel \\
	Fraunhofer AISEC \\
	\texttt{carla.sagebiel\footnotemark[\value{footnote}]} \\
	\And
	Konstantin B\"ottinger \\
	Fraunhofer AISEC \\
	\texttt{konstantin.boettinger\footnotemark[\value{footnote}]} \\
}

\maketitle 

\begin{abstract}
Neural networks follow a gradient-based learning scheme, adapting their mapping parameters by back-propagating the output loss.
Samples unlike the ones seen during training cause a different gradient distribution.
Based on this intuition, we design a novel semi-supervised anomaly detection method called \GAA{}.
By analysing the temporal distribution of the gradient over multiple training steps, we reliably detect point anomalies in strict semi-supervised settings.
Instead of domain dependent features, we input the raw gradient caused by the sample under test to an end-to-end recurrent neural network architecture.
\GAA{} works in a purely data-driven way, thus is readily applicable in a variety of important use cases of anomaly detection.

\end{abstract}

\section{Introduction}
Anomalies are inputs that significantly deviate from the given notion of normal.
Depending on the use case, anomalies may lead to attacks on the infrastructure, fraudulent transactions or points of interest in general.
In recent years, research on semi-supervised anomaly detection (AD) gained traction (e.g.\ \cite{ruff_deep_2020,pang_deep_2019,sperl_activation_2021}), where we leverage prior knowledge about the anomalous distribution to boost the overall detection performance.
This setting is often found in real-world settings, where a few anomalies have already been detected manually while the rest are unknown.
Unlike classification tasks, a semi-supervised AD method should not just differentiate between normal inputs and known anomalies, but also reveal yet unseen types of anomalies.

The lack of absolute training data modelling all types of anomalies complicates the use of machine learning algorithms with an automatic feature selection, e.g.\ deep learning (DL) methods.
In our research, we alleviate this problem by analysing an abstract representation of the input: its temporal gradient distribution.
Intuitively, a neural network (NN) trained only on the known normal data will fail to process anomalies in the same manner.
We analyse this discrepancy with the help of an auxiliary NN.
To reduce the manual work and domain expert knowledge required, we designed our AD method to be purely data-driven.
Instead of hand-crafted features, we analyse the raw gradient caused by individual inputs for anomalous patterns.
In our thorough empirical study, we show that our method generalises to several use cases and data types.
Based on this principle, we call our novel AD method \GAA{}, \underline{r}aw g\underline{rad}ient \underline{a}nomaly \underline{d}etection.
In summary, our contributions to AD research are:
\begin{itemize}
    \item We introduce a novel data-driven end-to-end neural architecture to analyse the temporal distribution of the gradient to detect point anomalies.
    \item To the best of our knowledge, \GAA{} is the first semi-supervised AD method based on the analysis of gradients.
    \item We thoroughly analyse the performance gain by \GAA{} on ten data sets against five baseline methods.
    \item To support future research, we plan to open-source our code.
\end{itemize}

\subsection{Related Work}
\GAA{} is a DL-based, semi-supervised AD method building on the analysis of the input's gradient space.
In the following, we discuss related work from all of the three categories.
For a broader overview on AD, we recommend the surveys of Pang~et~al. \cite{pang_deep_2021} and Ruff~et~al. \cite{ruff_unifying_2021}.

\paragraph{Anomaly Detection based on Deep Learning Methods}
DL methods deliver high performance even on complex inputs, but are data-demanding.
Due to the inherent class imbalance of AD, it is challenging to apply DL methods.
Over the past years, a variety of solutions arose, which we loosely group in three categories: methods based on 1) the reconstruction error, 2) the distance to the training data and 3) end-to-end architectures.
Reconstruction-based methods use a representation or distribution estimation method, e.g.\ autoencoders (AEs) \cite{borghesi_anomaly_2019,vu_anomaly_2019,beggel_robust_2020} or generative adversarial nets (GANs) \cite{schlegl_f-anogan_2019,li_mad-gan_2019,akcay_ganomaly_2019}.
Intuitively, when the network is fitted on the normal data, there is a measurable difference between the reconstructed and the input sample when an anomaly is processed.
The main problems are noisy data sets, causing a low reconstruction error for some anomalies, and anomalies close to normal samples, which are easy to reconstruct.
Distance-based methods, e.g.\ one-class classifiers \cite{chalapathy_deep_2019,ruff_deep_2020,sohn_learning_2020}, introduce a transformer network.
Using a suitable metric, the transformer network maps normal samples close to each other, but anomalies far away.
Problems may arise when the data set contains multiple notions of normal, which cannot be mapped to the very same centre of normality.
\GAA{} uses an end-to-end neural architecture, directly mapping the input to an anomaly score.
Usually, end-to-end architectures \cite{pang_learning_2018-1,pang_deep_2019,sperl_activation_2021} require normal as well as anomalous training samples.
However, manually finding anomalies is a time-consuming and error-prone process.
Research on substituting real anomalies by artificially created ones, e.g.\ geometric transformations \cite{golan_deep_2018,bergman_classification-based_2020} or out-of-distribution (OOD) samples \cite{hendrycks_deep_2019}, tries to solve this issue.
These methods need careful adaptations to the respective data set.
In \GAA{}, we mitigate the problem by using a simple source for trivial anomalies: a Gaussian distribution as done in \AAA{} \cite{sperl_activation_2021}.
Our evaluation motivates that our analysis in the gradient space of NNs allows to find a suitable boundary between real normal and real anomalous samples even with this simple source for counterexamples.

\paragraph{Semi-supervised Anomaly Detection}
In the past years, research about semi-supervised AD has gained traction.
In real-world scenarios, a few known anomalies -- much less than the normal samples -- may already be available.
These known anomalies may have been found manually or by an unsupervised AD method.
Semi-supervised AD methods use this kind of prior knowledge to boost the overall detection performance.
DeepSAD \cite{ruff_deep_2020} is a semi-supervised extension of one-class classifiers.
Deviation Networks (DevNet) \cite{pang_deep_2019} is based on distance metrics.
The authors of \AAA{} \cite{sperl_activation_2021} analyse the hidden activations of NNs for anomalous patterns.
Reconstruction errors are evaluated in ABC \cite{yamanaka_autoencoding_2019} and ESAD \cite{huang_esad_2021}.
Expanding the view to OOD detection, DROCC \cite{goyal_drocc_2020-1} uses generated counterexamples based on the prior knowledge about real anomalies.
For semi-supervised AD methods, the distribution of the known anomalies may severely impact the generalisation performance \cite{ye_understanding_2021}.
Thus, a main challenge is the detection of unknown anomalies, i.e.\ anomalies, which have not yet been detected manually.

\begin{table}[tb]
    \caption{
    Related work across different detection domains analysing the gradient space of NNs.
    }
    \label{tab:relatedwork}
    \rowcolors{2}{white}{gray!10}
    \centering
    \begin{tabular}{l >{\centering}p{2.75cm} >{\centering}p{2.75cm} >{\centering\arraybackslash}p{2.75cm}}
    \toprule
    & \bfseries Unsupervised & \bfseries Semi-supervised & \bfseries Supervised \\
    \midrule
    \bfseries Anomaly Det. & \cite{kwon_novelty_2020,kwon_backpropagated_2020} & \bfseries \GAA{} & -- \\
    \bfseries Out-of-Distr. Det. & \cite{huang_importance_2021,sun_gradient-based_2021} & \cite{lee_open-set_2021} & -- \\
    \bfseries Adversarial Det. & -- & -- & \cite{dhaliwal_gradient_2018,lust_gran_2020,schulze_da3g_2021} \\
    \bottomrule
    \end{tabular}
\end{table}

\paragraph{Gradient-based Detection of Anomalous Instances}
\GAA{} analyses the gradient space of NNs.
Despite the variety of AD research, this idea has barely been covered by previous work.
We give an overview in \Cref{tab:relatedwork}.
Kwon~et~al.\ \cite{kwon_novelty_2020} propose using the $l_2$-norm of an AE's gradient.
The same authors refine the idea in their AD method GradCon \cite{kwon_backpropagated_2020}.
Here, they measure the cosine similarity between past normal gradients and the current input.
Expanding the view to research topics related to AD, we see applications in OOD and adversarial detection.
In OOD detection, multi-class data and thus known class labels are assumed, which is not applicable to AD, where we merely distinguish between monolithic sets of normal and anomalous data.
Sun~et~al.\ detect OOD samples by measuring the Mahalanobis distance of the gradient.
In GradNorm \cite{huang_importance_2021}, the authors used the Kullback-Leibler divergence on the $l_1$-norm of the gradient.
Similarly, Lee~et~al.\ \cite{lee_open-set_2021} use the $l_1$-norm, but also incorporate some known OOD samples.
In adversarial detection, samples, which have been specifically generated to alter the decision of a NN, are detected.
In contrast to AD and OOD detection, adversarial detection is usually considered a supervised problem because counterexamples can be easily generated.
In GraN \cite{lust_gran_2020}, the authors used the $l_1$-norm of the gradient, whereas in Gradient Similarity \cite{dhaliwal_gradient_2018}, the authors took the $l_2$-norm of the gradient along the cosine similarity to distinguish between benign and adversarial samples.
In DA3G \cite{schulze_da3g_2021}, the authors analyse the raw gradient of the last two layers of classifiers.
In \GAA{}, we refrain from using hand-crafted features or manually selecting points of interest as each choice incorporates prior knowledge from the algorithm designer, which may not be backed by the training data.
Instead, we analyse the temporal distribution of the entire raw gradient by our end-to-end DL-based architecture.
Our evaluation shows that \GAA{} outperforms past AD methods on a variety of use cases and data types.

\section{Prerequisites}
In AD, we discover samples that deviate from the training data set $\set{X}_\text{norm}$.
Implicitly, we assume all samples in $\set{X}_\text{norm}$ to be normal, even when polluted by unknown anomalies.
In literature, there is some ambiguity in the definition of semi-supervised AD, which is sometimes referred to as supervised AD.
In this regard, we follow the notation of Ruff~et~al.\ \cite{ruff_deep_2020}.
In our semi-supervised scenario, further we have access to a small data set $\set{X}_\text{anom}$, containing a few known anomalies, i.e.\ $\abs{\set{X}_\text{norm}} \gg \abs{\set{X}_\text{anom}}$.
Note that AD differs from related topics as OOD detection.
In OOD detection, we do have access to an underlying classifier and its multi-class training data set.
Instead, in AD, we consider the entire normal data set as one class and detect deviations from it.
We refer to the survey of Salehi~et~al. \cite{salehi_unified_2021} for an in-depth discussion of AD and its related research topics.

\subsection{Activation Anomaly Analysis}
Parts of \GAA{} are inspired by the semi-supervised AD method \AAA{} \cite{sperl_activation_2021}.
Sperl et~al. introduce their so-called target-alarm architecture.
The target network, e.g. an AE, learns the distribution of the normal data.
An auxiliary NN, called the alarm network, analyses the hidden activations of the target network while processing normal as well as anomalous inputs.
As additional source of anomalous patterns, they input synthetic anomalies generated from a Gaussian prior.

In \GAA{}, we extend the target-alarm architecture to analyse the temporal gradient distribution of AEs.
We use a recurrent alarm network to concurrently analyse the gradient of multiple AEs for anomalous patterns.
Each AE reflects a different training state of the very same architecture.
Our evaluation shows that the temporal gradient distribution allows a more reliable anomaly detection performance even under severe data pollution and unknown anomalies.

\section{\GAA{}}

\begin{figure}[tb]
    \centering
\definecolor{f-basic}{RGB}{66, 69, 79} 
\definecolor{f-alarm}{RGB}{186, 37, 60} 
\definecolor{f-blue}{RGB}{35, 96, 194}
\definecolor{f-green}{RGB}{52, 166, 35}

\pgfdeclarelayer{bg}    
\pgfdeclarelayer{shapes}    
\pgfdeclarelayer{box}    
\pgfsetlayers{box, bg, shapes, main}  

\tikzstyle{rect} = [rectangle, minimum width= 1 cm, minimum height= 1 cm, text centered, node distance = 2 cm]
\tikzstyle{block} = [rect, text = white, minimum width= 1 cm, text width= 0.5 cm, minimum height= 1cm, node distance = 2cm ]
\tikzstyle{input} = [rect, minimum width = 0cm]
\tikzstyle{output} = [rect, minimum width = 0cm]
\tikzstyle{midres} = [rect, minimum width = 0.7cm, minimum height = 0.7 cm, text width = 0.7 cm, fill = white]
\tikzstyle{alarm} = [rect, fill = f-alarm, text = white, ]
\tikzstyle{loss} = [draw, rect, 
	minimum width = 1cm ]
\tikzstyle{empty} = [rect, fill = cyan!20, minimum width = 0.5cm, minimum height = 0.3 cm ]

\tikzstyle{line} = [-, shorten >=2pt]
\tikzstyle{arrow} = [line, ->]
\tikzstyle{sqarrow} = [arrow, decorate,  line join=round, decoration={
    snake,
    segment length=6,
    amplitude=1.2,post=lineto,
    post length=2pt]}]
 
\tikzstyle{triang} = [trapezium, minimum width=1.2cm, fill = f-basic, node distance = 0.25cm, minimum height = 0.5cm, shape border rotate = 270]
\newcommand{\autoencoder}[1]{
\begin{scope}[]
	\node [at = (#1)] (t){};
    	\node [triang, left of = t ] {};
	\node [triang, shape border rotate = 90,  right of = t ] () {}; 
\end{scope}
}

\begin{tikzpicture}

	\node [block] (fae0) {};
	\node [block, below of = fae0, node distance = 1.75 cm] (fae1) 
		{}; 
	\node [block, below of = fae1, node distance = 1.75 cm] (fae2) 
		{};
	\node [empty, text = black, text width = 1cm] 
	(fae11)  at ($(fae1)!0.5!(fae2)$)
	{
		\dots
	};
	
	\node [midres, right of = fae0] (xs0) {$\hat{\vect{x}}^{(t_{1})} $};
	\node [midres, right of = fae1] (xs1) {$\hat{\vect{x}}^{(t_{2})} $};
	\node [midres, right of = fae2] (xs2) {$\hat{\vect{x}}^{(t_{n})} $};
	
	\node [loss, right of = xs0] (loss0) {$\nabla \set{L} (\vect{x}, \hat{\vect{x}}^{(t_{1})}) $};
	\node [loss, right of = xs1] (loss1) {$\nabla \set{L} (\vect{x}, \hat{\vect{x}}^{(t_{2})}) $};
	\node [loss, right of = xs2] (loss2) {$\nabla \set{L} (\vect{x}, \hat{\vect{x}}^{(t_{n})}) $};
	
   	\node [alarm,  
		right of = loss0, node distance = 2.75cm] (alarm) {RNN};
	\node [alarm,  
		right of = loss1, node distance = 2.75cm] (alarm1) {RNN};
	\node [alarm,  
		right of = loss2, node distance = 2.75cm] (alarm2) {RNN};
	\node[alarm, right of = alarm2, node distance = 1.4cm](mapping){NN};
	\node [output, right of = mapping, text width = 0.3cm, node distance = 1.45cm] (output) {$\hat y$};
		
	\node [input, left of = mapping, node distance = 9.5cm, text width = 0.3cm] (input1) 
	{$\vect{x}$}
	;	
	
	\begin{pgfonlayer}{shapes}
		\autoencoder{fae0};
		\autoencoder{fae1};
		\autoencoder{fae2};
	\end{pgfonlayer}
	
	\begin{pgfonlayer}{bg}
		\draw [arrow] (input1.east) -- ($(input1)!0.4!(fae2)$) |- (fae0);
		\draw [arrow] (input1.east) -- ($(input1)!0.4!(fae2)$) |- (fae1);
		\draw [arrow] (input1.east) -- ($(input1)!0.4!(fae2)$) |- (fae2);
		
		\draw [arrow, cyan, dashed, very thick, -stealth] ($(fae0) + (0, 0.7)$) -- 
		node [below right, annot, pos = 0.22, cyan, text width=2.5cm] {Retrain for $T$ epochs}
		node [below right, annot, pos = 0.6, cyan, text width=2.5cm] {Retrain for $T$ epochs}
		($(fae2) + (0,-0.7)$);
		
		\draw [arrow] (fae0) -- (xs0);
		\draw [arrow] (fae1) -- (xs1);
		\draw [arrow] (fae2) -- (xs2);
		
		\draw [sqarrow] (xs0) -- (loss0);
		\draw [sqarrow] (xs1) -- (loss1);
		\draw [sqarrow] (xs2) -- (loss2);
		
		\draw [arrow] (loss0) -- node[annot, f-alarm] {Batch norm.} (alarm);
		\draw [arrow] (loss1) -- node[annot, f-alarm] {Batch norm.} (alarm1);
		\draw [arrow] (loss2) -- node[annot, f-alarm] {Batch norm.}  (alarm2);
		
		\draw [arrow] (alarm.east) -- ($(alarm.east) + (0.2,0)$) 
			|- ($0.5*(alarm) + 0.5*(alarm1)$)
			-|  ($(alarm1.west) + (-0.2,0.2)$) -- ($(alarm1.west) + (0,0.2)$);
		\draw [arrow] (alarm1.east) -- ($(alarm1.east) + (0.2,0)$) 
			|- ($0.5*(alarm1) + 0.5*(alarm2)$)node [fill = red!20] {$\dots$}
			-|  ($(alarm2.west) + (-0.2,0.2)$) -- ($(alarm2.west) + (0,0.2)$);	
		\draw [arrow] (alarm2) -- (output);
	\end{pgfonlayer}

	\begin{pgfonlayer}{box}
		\fill [fill=cyan!20 ] ($(fae0) + (-1, .75) $) -|  ($(fae2) + (1, -.75)$) -| 
		cycle ;
		
		\fill [fill=red!20 ] ($(alarm) + (-1, .75) $) -|  ($(mapping) + (1, -.75)$) -|  cycle ;
		\node [annot, below of = fae2, text width = 2cm, node distance = 1.3cm ] (title0) {Family of \\ Autoencoders} ;
		\node [annot, right of = title0, xshift=2.0cm, text width = 3cm, node distance = 1.7cm ] (title1) {Temporal Gradient\\Distribution} ;
		\node [annot, right of = title1, xshift=2.0cm, text width = 3cm, node distance = 1.7cm ] (title2) {Recurrent Alarm Network} ;
	\end{pgfonlayer}

\end{tikzpicture}
    \caption{
    Data flow of \GAA{}: we map the input sample $\vect{x}$ to an anomaly score $\hat{y} \in [0,1]$, where 1~is highly anomalous.
    The input is processed by a family of AEs, each yielded by successive training on the normal samples.
    We measure the discrepancy between the predicted and the original input by calculating the respective gradient.
    An auxiliary network, called the alarm network, analyses this sequence of gradients for anomalous patterns.
    }
    \label{fig:overview}
\end{figure}
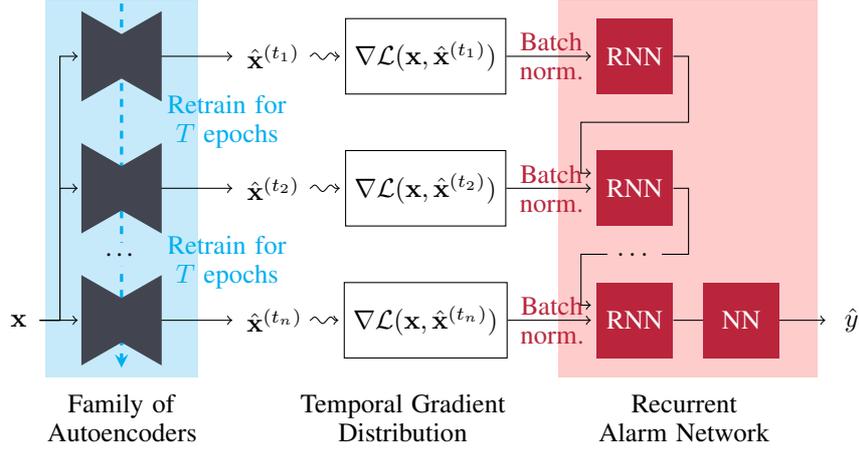

\GAA{} builds upon our main intuition:
\begin{quote}
    Let $f_\text{AE}(\vect{x}; \vectg{\theta})$ be an AE trained on the data set $\set{X}_\text{norm}$ containing normal samples.
    The temporal evolution of the raw gradient $\nabla f_\text{AE}(\vect{x})=\nabla_{\vectg{\theta}} \set{L} (\vect{x}, f_\text{AE}(\vect{x}; \vectg{\theta}))$ is useful to decide if the current input $\vect{x}$ is normal or anomalous.
\end{quote}

Let $f_\text{AE}^{(i)}(\vect{x}) = f_\text{AE}(\vect{x}; \vectg{\theta}^{(i)})$ be the target AE after the $i$-th training epoch.
With each training step, the mapping parameters $\vectg{\theta}$ adapt more to the training data.
According to our intuition, we analyse the temporal distribution of the gradient.
Let $g_{f_\text{AE}}(\vect{x})$ denote the function that extracts the gradients over time given the target AE $f_\text{AE}(\cdot)$:
\begin{equation}
\label{eq:gradient}
    g_{f_\text{AE}} (\vect{x}) = [\nabla f_\text{AE}^{(i)} (\vect{x})]_{i=T_0+jT, j\in\Naturals} = [\nabla f_\text{AE}^{(T_0)} (\vect{x}), \nabla f_\text{AE}^{(T_0+T)} (\vect{x}),  \ldots],
\end{equation}
where $T$ is some sampling frequency and $T_0$ an offset.

Given the temporal gradient distribution, an auxiliary NN, called the \emph{alarm network} $f_\text{alarm}(\cdot)$, analyses it for anomalous patterns.
The alarm network is a binary classifier outputting an anomaly score, where~1 is highly anomalous.
Both networks are combined to the overall end-to-end architecture of \GAA{} depicted in \Cref{fig:overview} and formally defined as:
\[
f_\text{\GAA{}}(\vect{x}) = f_\text{alarm}(g_{f_\text{AE}} (\vect{x})) \in [0, 1].
\]
Due to the sequential nature of the gradient, the alarm network is a recurrent neural network (RNN).
We combine the RNN with a time-distributed batch normalisation \cite{ioffe_batch_2015} layer and fully-connected output layers.
In our research, we found the batch normalisation layer to be essential to scale small gradients, especially after several training epochs of the target network.

\begin{algorithm}[tb]
\caption{High-level overview about \GAA{}'s training objectives.}\label{alg:training}
\KwIn{$f_\text{AE}(\vect{x}; \vectg{\theta}^{T_0}), \set{D}_\text{train} = \set{X}_\text{train} \times \set{Y}_\text{train}$}
\KwResult{$f_\text{\GAA{}}$}
\Comment{Retrain the autoencoder on $\set{X}_\text{norm} \subset \set{X}_\text{train}$}
$f_\text{AE,0} \gets f_\text{AE}(\vect{x}; \vectg{\theta}^{T_0})$,
$f_\text{AE,1} \gets f_\text{AE}(\vect{x}; \vectg{\theta}^{T_0+T})$,
$\ldots$\;

\For{$(\vect{x}, y) \in \set{X}_\text{train}$}{
    \Comment{Sample synthetic anomalies}
    $\tilde{\vect{x}} \gets \set{N}(\vect{.5}, \vect{1})$\;
    \Comment{Extract the gradients from the retrained autoencoders}
    $\vect{g} \gets [\nabla f_\text{AE,0} ( \vect{x}), \nabla f_\text{AE,1} ( \vect{x}), \ldots]$, 
    $\tilde{\vect{g}} \gets [\nabla f_\text{AE,0} ( \tilde{\vect{x}}), \nabla f_\text{AE,1} ( \tilde{\vect{x}}), \ldots]$, \cref{eq:gradient}\;
    \Comment{Train \GAA{}'s components}
    $\argmin \vectg{\theta}_\text{batchnorm.}: f_\text{alarm} \gets (\vect{g}, y)$, \cref{eq:batchnorm}\;
    $\argmin \vectg{\theta}_\text{alarm}: f_\text{alarm} \gets (\vect{g}, y),(\tilde{\vect{g}}, 1)$, \cref{eq:alarm}\;
}
\end{algorithm}

\subsubsection{Training Objectives}
AD is characterised by its inherent class imbalance, where known anomalies are rare and might not cover the entire anomaly distribution.
In \GAA{}, we solve this problem by sampling trivial counterexamples from a Gaussian prior, i.e. $\tilde{\vect{x}} \sim \set{N}(\mu, \sigma^2)$.
Even though these synthetic anomalies do not resemble real ones, our analysis in the gradient space results in a meaningful decision barrier between real normal and real anomalous inputs.
As result, the training objective of \GAA{} becomes a simple classification using the binary cross entropy (BXE) as loss:
\begin{equation}
\label{eq:alarm}
        \argmin_{\theta_\text{alarm}} \Expected [
\set{L}_\text{BXE} \left( y, f_\text{\GAA{}} \left( \vect{x} \right) \right)
+ \set{L}_\text{BXE} \left( 1, f_\text{\GAA{}} \left( \tilde{\vect{x}} \right) \right)
],
\end{equation}
where $(\vect{x},y) \sim P_\set{D}, \tilde{\vect{x}} \sim \set{N}( \vect{0.5}, \vect{1} )$.
Our input data is scaled to $\vect{x} \in [0,1]^N$, thus the synthetic anomalies are likely outside this interval, i.e. clearly anomalous.
Due to the random nature of the counterexamples, we adapt the batch normalisation layer on the training data only, i.e.:
\begin{equation}
\label{eq:batchnorm}
        \argmin_{\theta_\text{batchnorm.}} \Expected [
\set{L}_\text{BXE} \left( y, f_\text{\GAA{}} \left( \vect{x} \right) \right)
].
\end{equation}
In Algorithm~\ref{alg:training}, we summarise \GAA{}'s training process.

\begin{table}[tb]
\caption{
Data sets under evaluation.
If more than one anomaly class was available, we evaluated \GAA{} on anomaly classes unknown during training in our transfer experiments.
}
\label{tab:datasets}
\centering
\begin{tabular}{c c c c c c}
\textbf{Data}   & \textbf{Normal}   & \textbf{Train Ano.} & $\subseteq$  & \textbf{Test Ano.} & \textbf{Encoder} \\
\midrule
CC \cite{pozzolo_calibrating_2015} & Normal & Anomalous & & Anomalous & 20, 10, 5 \\
CoverType \cite{blackard_comparative_1999} & 1-3 & 4-5 &  & 4-7 & 40, 20, 10 \\
DarkNet \cite{habibi_lashkari_didarknet_2020} & Non-Tor/-VPN & Tor & & Tor, VPN & 60, 30, 15 \\
DoH \cite{montazerishatoori_detection_2020} & Benign & Mal. & & Mal. & 20, 10, 5 \\
FMNIST \cite{xiao_fashion-mnist_2017} & 0-3 & 4-6 & & 6-9 & 8C3-8C3-8 \\
IDS \cite{sharafaldin_toward_2018} & Benign & Bot, BF & & Bot, BF, Infil., Web & 60, 40, 20 \\
KDD \cite{tavallaee_detailed_2009} & Normal & DoS, Probe & & DoS, Probe, R2L, U2R & 40, 20, 10 \\
MNIST \cite{lecun_gradient-based_1998} & 0-3 & 4-6 & & 6-9 & 8C3-8C3-8 \\
Mam. \cite{woods_comparative_1993} & Normal &  Malignant & & Malignant & 5, 3, 2 \\
URL \cite{mamun_detecting_2016} & Benign & Def., Mal. & & Def., Mal., Phi., Spam & 60, 30, 15 \\
\end{tabular}
\end{table}

\section{Experimental Setup}
\label{sec:setup}
We evaluated \GAA{} in challenging experiments mimicking real-world scenarios.
In \Cref{tab:datasets}, we show the ten data sets under evaluation, ranging from commonly used baseline data sets to important applications of AD, e.g. intrusion or fraud detection.
We scaled all numerical values to $[0,1]$ and 1-Hot encoded categorical entries.
If not given by the data set, 75\% were used for the training split, 5\% for validation and 20\% for testing.
While training \GAA{}'s AE, 25\% of the training data were held back to evaluate the gradient distribution of some fresh normal samples while training the alarm network.

\subsubsection{Baseline Methods}
\GAA{} is a deep semi-supervised AD method based on the analysis of the gradient space of AEs.
AEs themselves can be used as AD method by measuring the reconstruction error, when only trained on the normal data.
We used the mean squared error as anomaly score, i.e. $\hat{y} = \norm{f_\text{AE}(\vect{x}) - \vect{x}}$.
GradCon \cite{kwon_backpropagated_2020} is a AD method based on the analysis of the gradient space of NNs.
We favoured GradCon over the authors' initial AD method based on $l_2$-norms \cite{kwon_novelty_2020} as it generally performed better according to their evaluation.
Both aforementioned baseline methods are unsupervised, thus do not profit from known anomalies.
Expanding our view to deep semi-supervised AD, DeepSAD \cite{ruff_deep_2020} is a commonly used baseline.
In the same category, DevNet \cite{pang_deep_2019} and \AAA{} \cite{sperl_activation_2021} are currently the best performing methods.
We compared \GAA{} to all of these challenging baselines to assess the performance increase by our research.

\subsubsection{Parameter Choices}
We designed \GAA{} as a data-driven method, which readily applies to a diverse set of use cases and data types.
Thus, we chose one common set of hyperparameters for the entire evaluation.
Across all data sets, we analysed a target network trained for $T_0=10$ epochs across 2~retraining steps, each with $T=5$ epochs resulting in three models.
The alarm network had the dimensions $1000, 500, 200, 75$ except for the small Mammography data set, where we used $100, 50, 25, 10$.
LSTM \cite{hochreiter_long_1997} elements were used for the first two dimensions, ReLU-activated dense layers else.
\GAA{} was trained for 100 epochs at a learning rate of $0.001$ using Adam as optimiser.
For a fair comparison, we chose the same hyperparameters for the other baseline methods if applicable.

\section{Evaluation}
In our evaluation, we compared \GAA{}'s detection performance to five challenging baseline methods across ten data sets.
We carefully followed the best practices introduced by Hendrycks~\&~Gimpel \cite{hendrycks_baseline_2017} and report the performance as area under the ROC curve (AUC) and average precision (AP).
Both metrics measure the performance independently of a chosen detection threshold.
An ideal AD method scores an AUC and AP of $1$.
To measure the significance of our results, we report the p-value of the Wilcoxon signed-rank test \cite{wilcoxon_individual_1992}.
It evaluates the null hypothesis that a ranked list of measurements was derived from the same distribution.

\begin{table}[tb]
    \caption{
    Detection of known anomalies, i.e. the training and test data set contained the same anomaly classes.
    We limited the number of known anomalies to 100 and show the results after five detection runs.
    }
    \label{tab:results:known}

\centering
    
\resizebox{\textwidth}{!}{%
\begin{tabular}{H H >{\color{gray}}c >{\color{gray}}c >{\color{gray}}c | >{\color{gray}}c >{\color{gray}}c >{\color{gray}}c >{\color{gray}}c | >{\color{gray}}c >{\color{gray}}c >{\color{gray}}c >{\color{gray}}c >{\color{gray}}c >{\color{gray}}c }
    & & & \multicolumn{2}{>{\color{gray}}c}{Ours} &                            \multicolumn{4}{>{\color{gray}}c}{Unsupervised Baselines} &                            \multicolumn{6}{>{\color{gray}}c}{Semi-supervised Baselines} \\

    & & & \multicolumn{2}{c}{\GAA{}} & \multicolumn{2}{c}{AE} & \multicolumn{2}{c}{GradCon} & \multicolumn{2}{c}{DeepSAD} & \multicolumn{2}{c}{DevNet} & \multicolumn{2}{c}{\AAA{}} \\
    & & &                                              AUC &                                               AP &                                              AUC &                                               AP &                                              AUC &                                               AP &                                 AUC &                                  AP &                                              AUC &                                               AP &                                              AUC &                                               AP \\
\midrule
 & & CC &  \color{black}$.98 \scriptscriptstyle \pm .01$ &  \color{black}$.81 \scriptscriptstyle \pm .03$ &               $.95 \scriptscriptstyle \pm .00$ &               $.43 \scriptscriptstyle \pm .00$ &  $.80 \scriptscriptstyle \pm .16$ &  $.34 \scriptscriptstyle \pm .08$ &  $.88 \scriptscriptstyle \pm .03$ &               $.34 \scriptscriptstyle \pm .26$ &  \color{black}$.98 \scriptscriptstyle \pm .00$ &               $.74 \scriptscriptstyle \pm .00$ &               $.88 \scriptscriptstyle \pm .06$ &               $.51 \scriptscriptstyle \pm .21$ \\
    &     & CT &  \color{black}$.84 \scriptscriptstyle \pm .02$ &  \color{black}$.43 \scriptscriptstyle \pm .05$ &               $.76 \scriptscriptstyle \pm .02$ &               $.25 \scriptscriptstyle \pm .03$ &  $.70 \scriptscriptstyle \pm .05$ &  $.21 \scriptscriptstyle \pm .06$ &  $.57 \scriptscriptstyle \pm .07$ &               $.16 \scriptscriptstyle \pm .05$ &               $.83 \scriptscriptstyle \pm .01$ &               $.33 \scriptscriptstyle \pm .02$ &               $.46 \scriptscriptstyle \pm .06$ &               $.08 \scriptscriptstyle \pm .01$ \\
    &     & DN &  \color{black}$.92 \scriptscriptstyle \pm .01$ &  \color{black}$.75 \scriptscriptstyle \pm .02$ &               $.54 \scriptscriptstyle \pm .01$ &               $.23 \scriptscriptstyle \pm .01$ &  $.62 \scriptscriptstyle \pm .09$ &  $.24 \scriptscriptstyle \pm .04$ &  $.68 \scriptscriptstyle \pm .15$ &               $.36 \scriptscriptstyle \pm .12$ &               $.90 \scriptscriptstyle \pm .01$ &               $.69 \scriptscriptstyle \pm .02$ &               $.84 \scriptscriptstyle \pm .02$ &               $.54 \scriptscriptstyle \pm .05$ \\
    &     & DoH &  \color{black}$.98 \scriptscriptstyle \pm .00$ &  \color{black}$1.00 \scriptscriptstyle \pm .00$ &               $.85 \scriptscriptstyle \pm .01$ &               $.98 \scriptscriptstyle \pm .00$ &  $.74 \scriptscriptstyle \pm .06$ &  $.97 \scriptscriptstyle \pm .01$ &  $.73 \scriptscriptstyle \pm .08$ &               $.97 \scriptscriptstyle \pm .01$ &               $.91 \scriptscriptstyle \pm .01$ &               $.99 \scriptscriptstyle \pm .00$ &               $.83 \scriptscriptstyle \pm .04$ &               $.98 \scriptscriptstyle \pm .01$ \\
    &     & FMN. &               $.92 \scriptscriptstyle \pm .00$ &               $.95 \scriptscriptstyle \pm .00$ &               $.86 \scriptscriptstyle \pm .00$ &               $.92 \scriptscriptstyle \pm .00$ &  $.82 \scriptscriptstyle \pm .02$ &  $.88 \scriptscriptstyle \pm .03$ &  $.69 \scriptscriptstyle \pm .02$ &               $.77 \scriptscriptstyle \pm .02$ &               $.93 \scriptscriptstyle \pm .02$ &               $.96 \scriptscriptstyle \pm .01$ &  \color{black}$.95 \scriptscriptstyle \pm .01$ &  \color{black}$.97 \scriptscriptstyle \pm .00$ \\
    &     & IDS &  \color{black}$.93 \scriptscriptstyle \pm .01$ &  \color{black}$.89 \scriptscriptstyle \pm .01$ &               $.84 \scriptscriptstyle \pm .01$ &               $.52 \scriptscriptstyle \pm .04$ &  $.45 \scriptscriptstyle \pm .10$ &  $.19 \scriptscriptstyle \pm .06$ &  $.67 \scriptscriptstyle \pm .08$ &               $.38 \scriptscriptstyle \pm .12$ &               $.87 \scriptscriptstyle \pm .01$ &               $.67 \scriptscriptstyle \pm .07$ &               $.88 \scriptscriptstyle \pm .02$ &               $.67 \scriptscriptstyle \pm .11$ \\
    &     & KDD &               $.88 \scriptscriptstyle \pm .02$ &               $.90 \scriptscriptstyle \pm .03$ &  \color{black}$.95 \scriptscriptstyle \pm .00$ &  \color{black}$.95 \scriptscriptstyle \pm .00$ &  $.74 \scriptscriptstyle \pm .04$ &  $.83 \scriptscriptstyle \pm .01$ &  $.85 \scriptscriptstyle \pm .07$ &               $.88 \scriptscriptstyle \pm .06$ &               $.92 \scriptscriptstyle \pm .03$ &               $.94 \scriptscriptstyle \pm .01$ &               $.93 \scriptscriptstyle \pm .04$ &  \color{black}$.95 \scriptscriptstyle \pm .02$ \\
    &     & MN. &               $.97 \scriptscriptstyle \pm .01$ &  \color{black}$.98 \scriptscriptstyle \pm .00$ &               $.75 \scriptscriptstyle \pm .01$ &               $.79 \scriptscriptstyle \pm .01$ &  $.82 \scriptscriptstyle \pm .04$ &  $.84 \scriptscriptstyle \pm .04$ &  $.70 \scriptscriptstyle \pm .02$ &               $.75 \scriptscriptstyle \pm .02$ &               $.97 \scriptscriptstyle \pm .00$ &  \color{black}$.98 \scriptscriptstyle \pm .00$ &  \color{black}$.98 \scriptscriptstyle \pm .00$ &  \color{black}$.98 \scriptscriptstyle \pm .01$ \\
    &     & Mam. &  \color{black}$.94 \scriptscriptstyle \pm .01$ &  \color{black}$.69 \scriptscriptstyle \pm .03$ &               $.90 \scriptscriptstyle \pm .01$ &               $.27 \scriptscriptstyle \pm .01$ &  $.89 \scriptscriptstyle \pm .01$ &  $.26 \scriptscriptstyle \pm .02$ &  $.69 \scriptscriptstyle \pm .23$ &               $.16 \scriptscriptstyle \pm .12$ &  \color{black}$.94 \scriptscriptstyle \pm .01$ &               $.65 \scriptscriptstyle \pm .04$ &               $.88 \scriptscriptstyle \pm .04$ &               $.43 \scriptscriptstyle \pm .06$ \\
    &     & URL &  \color{black}$.95 \scriptscriptstyle \pm .01$ &  \color{black}$.99 \scriptscriptstyle \pm .00$ &               $.92 \scriptscriptstyle \pm .00$ &               $.98 \scriptscriptstyle \pm .00$ &  $.90 \scriptscriptstyle \pm .01$ &  $.97 \scriptscriptstyle \pm .00$ &  $.94 \scriptscriptstyle \pm .01$ &  \color{black}$.99 \scriptscriptstyle \pm .00$ &  \color{black}$.95 \scriptscriptstyle \pm .01$ &  \color{black}$.99 \scriptscriptstyle \pm .00$ &               $.94 \scriptscriptstyle \pm .01$ &  \color{black}$.99 \scriptscriptstyle \pm .00$ \\
    &     & mean &                              \color{black}$.93$ &                              \color{black}$.84$ &                                           $.83$ &                                           $.63$ &                              $.75$ &                              $.57$ &                              $.74$ &                                           $.58$ &                                           $.92$ &                                           $.79$ &                                           $.86$ &                                           $.71$ \\
    &     & p-val &                                                - &                                                - &                                           $.02$ &                                           $.01$ &                              $.00$ &                              $.00$ &                              $.00$ &                                           $.00$ &                                           $.38$ &                                           $.05$ &                                           $.06$ &                                           $.05$ \\

\end{tabular}
}

\end{table}

\subsection{Known Anomalies}
\label{sec:eval:known}
In our first experiment, we evaluated the performance gain in an ideal semi-supervised AD setting.
We limited the number of known anomalies to 100 randomly chosen samples, i.e. far less than normal samples available.
In \Cref{tab:results:known}, we summarise the results.
\GAA{} took the lead across all baseline methods, scoring the best on 7 out of 10 data sets.

As expected, the unsupervised baseline methods could not match the performance of the semi-supervised methods as they do not profit from the known anomalies.
Looking at the AUC, \GAA{} was \percincrease{.75}{.93}better than the other gradient-based AD method, GradCon.
KDD was the only data set, where the unsupervised methods took the lead.
Here, some unknown anomalies are within the test data set.
Similar to the discussion of Ye~et~al. \cite{ye_understanding_2021}, we believe the semi-supervised methods overfitted to the known anomalies.
Comparing our performance to GradCon, we see strong evidence that the analysis of the raw gradient is favourable over a hand-crafted feature set:
GradCon's analysis of the cosine similarity works well on some data sets (e.g. MNIST and DarkNet), but does not generalise to all ten data sets.
\GAA{} had the more consistent performance.

Considering the semi-supervised baselines, the largest margin was on DoH, where \GAA{} performed \percincrease{.91}{.98}better than DevNet, and \percincrease{.88}{.93}better on IDS compared to \AAA{}.
\AAA{} has a similar architecture as \GAA{}, but analyses the hidden activations of a single AE instead of the temporal gradient distribution.
Overall \GAA{} performed \percincrease{.86}{.93}better than \AAA{}.
Only on the image data sets, \AAA{} was the preferable method.
Summarising this section, \GAA{} clearly profited from the prior knowledge available in semi-supervised AD and allowed a more reliable detection performance compared to other state-of-the-art methods.

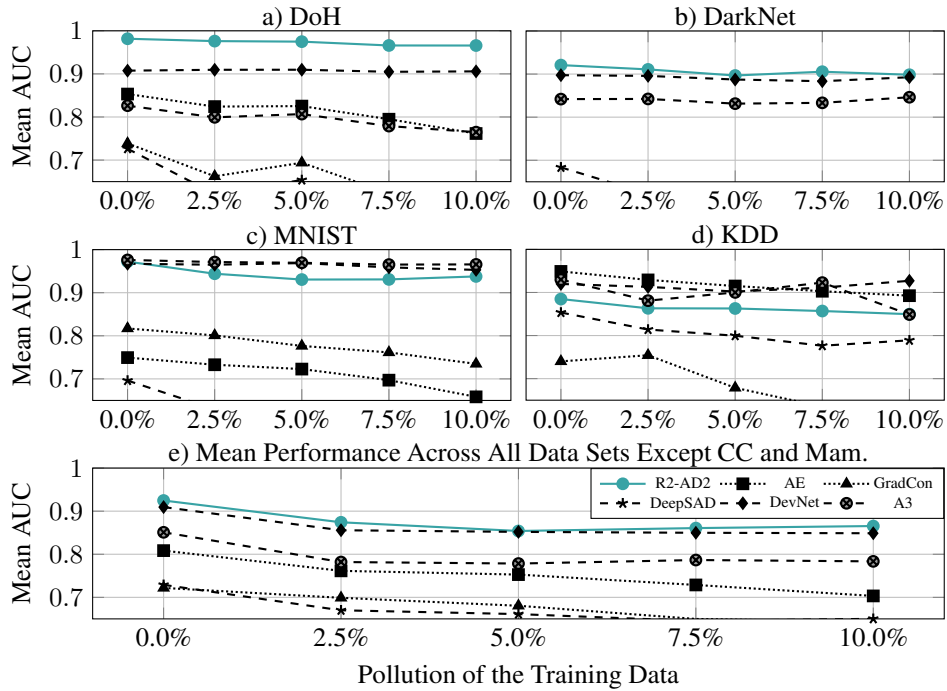
\begin{figure}[tb]
  \begin{center}
    \pgfplotstableread[col sep=comma]{csv/poll_mean.csv}\mean
\pgfplotstableread[col sep=comma]{csv/poll_doh.csv}\doh
\pgfplotstableread[col sep=comma]{csv/poll_darknet.csv}\darknet
\pgfplotstableread[col sep=comma]{csv/poll_mnist.csv}\mnist
\pgfplotstableread[col sep=comma]{csv/poll_kdd.csv}\kdd

\begin{tikzpicture}

\pgfmathsetmacro\plotw{5.5cm} 
\pgfmathsetmacro\ploth{2.0cm} 
\pgfmathsetmacro\vsep{.9cm}  
\pgfmathsetmacro\hsep{.2cm}  

	\begin{groupplot}[
		group style={
			group size=2 by 3,
			vertical sep=\vsep, horizontal sep=\hsep,
			xlabels at=edge bottom,
			ylabels at=edge left,
			y descriptions at=edge left,
		},
		legend style={
			at={(1, 1)},anchor=north east, legend columns=3, nodes={scale=0.6, transform shape}
		},
		title style={yshift=-2ex},
		y label style={yshift=-.0cm},
		height=\ploth, width=\plotw,
		grid=both,
		scale only axis,
		xtick=data,
		cycle list name=mark list,
		ymin=.65, ymax=1.0,
		symbolic x coords={{0.0}\%,{2.5}\%,{5.0}\%,{7.5}\%,{10.0}\%},
		xlabel={Pollution of the Training Data},
		ylabel={Mean AUC},
	]
		\nextgroupplot[title={a) DoH}]
		\addplot+[ad_line, error_p] table[x=Pollution,y=GAA-3-AUC-mean,y error=GAA-3-AUC-std] {\doh};
		\addplot+[unsu_line, error_p] table[x=Pollution,y=AE-AUC-mean,y error=AE-AUC-std] {\doh};
		\addplot+[unsu_line, error_p] table[x=Pollution,y=GradCon-AUC-mean,y error=GradCon-AUC-std] {\doh};
		\addplot+[semi_line, error_p] table[x=Pollution,y=DeepSAD-AUC-mean,y error=DeepSAD-AUC-std] {\doh};
		\addplot+[semi_line, error_p] table[x=Pollution,y=DevNet-AUC-mean,y error=DevNet-AUC-std] {\doh};
		\addplot+[semi_line, error_p] table[x=Pollution,y=A3-AUC-mean,y error=A3-AUC-std] {\doh};

		\nextgroupplot[title={b) DarkNet}]
		\addplot+[ad_line] table[x=Pollution,y=GAA-3-AUC-mean] {\darknet};
		\addplot+[unsu_line, error_p] table[x=Pollution,y=AE-AUC-mean,y error=AE-AUC-std] {\darknet};
		\addplot+[unsu_line, error_p] table[x=Pollution,y=GradCon-AUC-mean,y error=GradCon-AUC-std] {\darknet};
		\addplot+[semi_line, error_p] table[x=Pollution,y=DeepSAD-AUC-mean,y error=DeepSAD-AUC-std] {\darknet};
		\addplot+[semi_line, error_p] table[x=Pollution,y=DevNet-AUC-mean,y error=DevNet-AUC-std] {\darknet};
		\addplot+[semi_line, error_p] table[x=Pollution,y=A3-AUC-mean,y error=A3-AUC-std] {\darknet};

		\nextgroupplot[title={c) MNIST}]
		\addplot+[ad_line] table[x=Pollution,y=GAA-3-AUC-mean] {\mnist};
		\addplot+[unsu_line, error_p] table[x=Pollution,y=AE-AUC-mean,y error=AE-AUC-std] {\mnist};
		\addplot+[unsu_line, error_p] table[x=Pollution,y=GradCon-AUC-mean,y error=GradCon-AUC-std] {\mnist};
		\addplot+[semi_line, error_p] table[x=Pollution,y=DeepSAD-AUC-mean,y error=DeepSAD-AUC-std] {\mnist};
		\addplot+[semi_line, error_p] table[x=Pollution,y=DevNet-AUC-mean,y error=DevNet-AUC-std] {\mnist};
		\addplot+[semi_line, error_p] table[x=Pollution,y=A3-AUC-mean,y error=A3-AUC-std] {\mnist};

		\nextgroupplot[title={d) KDD}]
		\addplot+[ad_line] table[x=Pollution,y=GAA-3-AUC-mean] {\kdd};
		\addplot+[unsu_line, error_p] table[x=Pollution,y=AE-AUC-mean,y error=AE-AUC-std] {\kdd};
		\addplot+[unsu_line, error_p] table[x=Pollution,y=GradCon-AUC-mean,y error=GradCon-AUC-std] {\kdd};
		\addplot+[semi_line, error_p] table[x=Pollution,y=DeepSAD-AUC-mean,y error=DeepSAD-AUC-std] {\kdd};
		\addplot+[semi_line, error_p] table[x=Pollution,y=DevNet-AUC-mean,y error=DevNet-AUC-std] {\kdd};
		\addplot+[semi_line, error_p] table[x=Pollution,y=A3-AUC-mean,y error=A3-AUC-std] {\kdd};

		\nextgroupplot[
			title={e) Mean Performance Across All Data Sets Except CC and Mam.},
			width=2*\plotw+\hsep, xshift=0.5*(\plotw+\hsep),
		]
		\addplot+[ad_line] table[x=Pollution,y=GAA-3-AUC-mean] {\mean};
		\addplot+[unsu_line] table[x=Pollution,y=AE-AUC-mean] {\mean};
		\addplot+[unsu_line] table[x=Pollution,y=GradCon-AUC-mean] {\mean};
		\addplot+[semi_line] table[x=Pollution,y=DeepSAD-AUC-mean] {\mean};
		\addplot+[semi_line] table[x=Pollution,y=DevNet-AUC-mean] {\mean};
		\addplot+[semi_line] table[x=Pollution,y=A3-AUC-mean] {\mean};

		\legend{R2-AD2, AE, GradCon, DeepSAD, DevNet, A3}
	\end{groupplot}
\end{tikzpicture}
    \caption{
        Detection performance depending on the training data pollution.
        All semi-supervised methods had access to 100 known anomalies.
        Note that CC and Mammography did not contain enough anomalies, thus were excluded.
    }
    \label{fig:pollution}
  \end{center}
\end{figure}

\subsection{Noise Resistance}
\label{sec:eval:noise}
In real-world settings, it is usually infeasible to guarantee a clean training data set.
We evaluated this scenario by polluting the data with anomalous training samples labelled as normal.
All semi-supervised methods still had access to 100 known anomalies.
We summarise the performance in \Cref{fig:pollution} for DoH and DarkNet, where \GAA{} took the lead in our first experiment, and MNIST and KDD, where the baseline methods performed better.
Additionally, we show the mean performance across all data sets, which scaled to this experiment.

Looking at the mean performance, \GAA{} took the lead across all pollution levels.
The performance dropped only by \percincrease{.925}{.865}, when every tenth training sample was an anomaly labelled as normal.
For the unsupervised baselines, the performance drop was considerably larger, e.g. \percincrease{.808}{.703} for the AE.
The known anomalies seemed to stabilise the performance.
Across the data sets, the general ranking between the baseline methods did not change: AD methods that performed well on cleaner data sets also performed well on polluted data sets.
Our evaluation showed that \GAA{} is resistant to noisy training data sets as often found in real-world settings.

\begin{figure}[tb]
  \begin{center}
    \pgfplotstableread[col sep=comma]{csv/n_mean.csv}\mean
\pgfplotstableread[col sep=comma]{csv/n_doh.csv}\doh
\pgfplotstableread[col sep=comma]{csv/n_darknet.csv}\darknet
\pgfplotstableread[col sep=comma]{csv/n_mnist.csv}\mnist
\pgfplotstableread[col sep=comma]{csv/n_kdd.csv}\kdd

\begin{tikzpicture}

\pgfmathsetmacro\plotw{5.5 cm} 
\pgfmathsetmacro\ploth{2.0 cm} 
\pgfmathsetmacro\vsep{.9cm}  
\pgfmathsetmacro\hsep{.2cm}  

	\begin{groupplot}[
		group style={
			group size=2 by 3,
			vertical sep=\vsep, horizontal sep=\hsep,
			xlabels at=edge bottom,
			ylabels at=edge left,
			y descriptions at=edge left,
		},
		legend style={
			at={(0, 1)},anchor=north west, legend columns=3, nodes={scale=0.6, transform shape}
		},
		title style={yshift=-2ex},
		y label style={yshift=-.0cm},
		height=\ploth, width=\plotw,
		grid=both,
		scale only axis,
		xtick=data,
		xmin=0, xmax=105,
		ymin=.65, ymax=1.0,
		cycle list name=mark list,
		xlabel={Number of Known Anomalies During Training},
		ylabel={Mean AUC},
	]
		\nextgroupplot[title={a) DoH}]
		\addplot+[ad_line, error_p] table[x=NAnomalies,y=GAA-3-AUC-mean,y error=GAA-3-AUC-std] {\doh};
		\addplot+[unsu_line, error_p] table[x=NAnomalies,y=AE-AUC-mean,y error=AE-AUC-std] {\doh};
		\addplot+[unsu_line, error_p] table[x=NAnomalies,y=GradCon-AUC-mean,y error=GradCon-AUC-std] {\doh};
		\addplot+[semi_line, error_p] table[x=NAnomalies,y=DeepSAD-AUC-mean,y error=DeepSAD-AUC-std] {\doh};
		\addplot+[semi_line, error_p] table[x=NAnomalies,y=DevNet-AUC-mean,y error=DevNet-AUC-std] {\doh};
		\addplot+[semi_line, error_p] table[x=NAnomalies,y=A3-AUC-mean,y error=A3-AUC-std] {\doh};

		\nextgroupplot[title={b) DarkNet}]
		\addplot+[ad_line] table[x=NAnomalies,y=GAA-3-AUC-mean] {\darknet};
		\addplot+[unsu_line, error_p] table[x=NAnomalies,y=AE-AUC-mean,y error=AE-AUC-std] {\darknet};
		\addplot+[unsu_line, error_p] table[x=NAnomalies,y=GradCon-AUC-mean,y error=GradCon-AUC-std] {\darknet};
		\addplot+[semi_line, error_p] table[x=NAnomalies,y=DeepSAD-AUC-mean,y error=DeepSAD-AUC-std] {\darknet};
		\addplot+[semi_line, error_p] table[x=NAnomalies,y=DevNet-AUC-mean,y error=DevNet-AUC-std] {\darknet};
		\addplot+[semi_line, error_p] table[x=NAnomalies,y=A3-AUC-mean,y error=A3-AUC-std] {\darknet};

		\nextgroupplot[title={c) MNIST}]
		\addplot+[ad_line] table[x=NAnomalies,y=GAA-3-AUC-mean] {\mnist};
		\addplot+[unsu_line, error_p] table[x=NAnomalies,y=AE-AUC-mean,y error=AE-AUC-std] {\mnist};
		\addplot+[unsu_line, error_p] table[x=NAnomalies,y=GradCon-AUC-mean,y error=GradCon-AUC-std] {\mnist};
		\addplot+[semi_line, error_p] table[x=NAnomalies,y=DeepSAD-AUC-mean,y error=DeepSAD-AUC-std] {\mnist};
		\addplot+[semi_line, error_p] table[x=NAnomalies,y=DevNet-AUC-mean,y error=DevNet-AUC-std] {\mnist};
		\addplot+[semi_line, error_p] table[x=NAnomalies,y=A3-AUC-mean,y error=A3-AUC-std] {\mnist};

		\nextgroupplot[title={d) KDD}]
		\addplot+[ad_line] table[x=NAnomalies,y=GAA-3-AUC-mean] {\kdd};
		\addplot+[unsu_line, error_p] table[x=NAnomalies,y=AE-AUC-mean,y error=AE-AUC-std] {\kdd};
		\addplot+[unsu_line, error_p] table[x=NAnomalies,y=GradCon-AUC-mean,y error=GradCon-AUC-std] {\kdd};
		\addplot+[semi_line, error_p] table[x=NAnomalies,y=DeepSAD-AUC-mean,y error=DeepSAD-AUC-std] {\kdd};
		\addplot+[semi_line, error_p] table[x=NAnomalies,y=DevNet-AUC-mean,y error=DevNet-AUC-std] {\kdd};
		\addplot+[semi_line, error_p] table[x=NAnomalies,y=A3-AUC-mean,y error=A3-AUC-std] {\kdd};

		\nextgroupplot[
			title={e) Mean Performance Across All Data Sets},
			width=2*\plotw+\hsep, xshift=0.5*(\plotw+\hsep),
		]
		\addplot+[ad_line] table[x=NAnomalies,y=GAA-3-AUC-mean] {\mean};
		\addplot+[unsu_line] table[x=NAnomalies,y=AE-AUC-mean] {\mean};
		\addplot+[unsu_line] table[x=NAnomalies,y=GradCon-AUC-mean] {\mean};
		\addplot+[semi_line] table[x=NAnomalies,y=DeepSAD-AUC-mean] {\mean};
		\addplot+[semi_line] table[x=NAnomalies,y=DevNet-AUC-mean] {\mean};
		\addplot+[semi_line] table[x=NAnomalies,y=A3-AUC-mean] {\mean};

		\legend{R2-AD2, AE, GradCon, DeepSAD, DevNet, A3}
	\end{groupplot}
\end{tikzpicture}
    \caption{
        Detection performance depending on the number of known anomalies during training.
        The training data was not polluted by unknown anomalies.
    }
    \label{fig:anomalies}
  \end{center}
\end{figure}
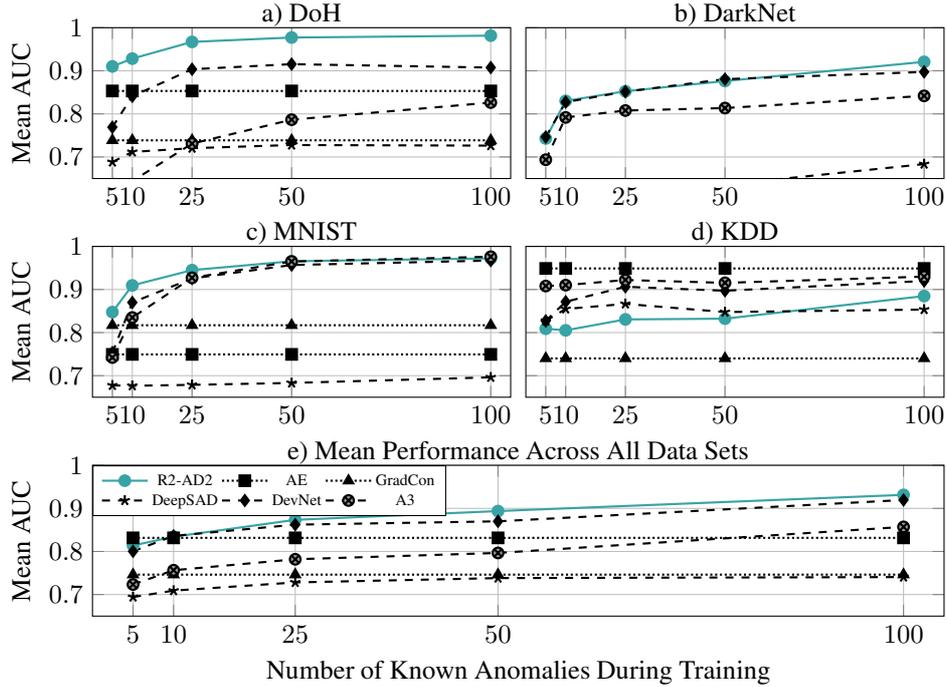

\subsection{Number of Known Anomalies}
\label{sec:eval:number}
In our next experiment, we evaluated the impact of the number of known anomalies available during training.
We gradually decreased the amount towards unsupervised regimes shown in \Cref{fig:anomalies}.
As the known anomalies were randomly selected among all anomaly classes, some classes might have been excluded during training.

As expected, the unsupervised methods remained at their initial performance as they do not incorporate known anomalies.
\GAA{} exceeded the performance of the AE with as little as ten known anomalies.
Looking at the mean performance, \GAA{} was better than the semi-supervised methods across all anomaly counts.
Interestingly, we saw \GAA{} to take the lead on MNIST for small amounts of prior knowledge.
For less than 50 anomalies, \GAA{} led on this data set.
In this experiment, we saw \GAA{} to perform well even with little prior knowledge about known anomalies. 

\begin{table}[tb]
    \caption{
    Detection of unknown anomalies, i.e. there are more anomaly classes in the test set than there were in the training data.
    We limited the number of known anomalies to 100 and show the results after five detection runs.
    }
    \label{tab:results:unknown}

\centering
    
\resizebox{\textwidth}{!}{%
\begin{tabular}{H H >{\color{gray}}c >{\color{gray}}c >{\color{gray}}c | >{\color{gray}}c >{\color{gray}}c >{\color{gray}}c >{\color{gray}}c | >{\color{gray}}c >{\color{gray}}c >{\color{gray}}c >{\color{gray}}c >{\color{gray}}c >{\color{gray}}c }
    & & & \multicolumn{2}{>{\color{gray}}c}{Ours} &                            \multicolumn{4}{>{\color{gray}}c}{Unsupervised Baselines} &                            \multicolumn{6}{>{\color{gray}}c}{Semi-supervised Baselines} \\

    & & & \multicolumn{2}{c}{\GAA{}} & \multicolumn{2}{c}{AE} & \multicolumn{2}{c}{GradCon} & \multicolumn{2}{c}{DeepSAD} & \multicolumn{2}{c}{DevNet} & \multicolumn{2}{c}{\AAA{}} \\
    & & &                                              AUC &                                               AP &                                              AUC &                                               AP &                                              AUC &                                               AP &                                 AUC &                                  AP &                                              AUC &                                               AP &                                              AUC &                                               AP \\

\midrule
& & CT &               $.64 \scriptscriptstyle \pm .02$ &               $.21 \scriptscriptstyle \pm .01$ &  \color{black}$.76 \scriptscriptstyle \pm .02$ &  \color{black}$.25 \scriptscriptstyle \pm .03$ &  $.70 \scriptscriptstyle \pm .05$ &  $.21 \scriptscriptstyle \pm .06$ &  $.51 \scriptscriptstyle \pm .07$ &  $.12 \scriptscriptstyle \pm .05$ &               $.63 \scriptscriptstyle \pm .05$ &               $.16 \scriptscriptstyle \pm .02$ &               $.38 \scriptscriptstyle \pm .07$ &               $.07 \scriptscriptstyle \pm .02$ \\
    &     & DN &  \color{black}$.86 \scriptscriptstyle \pm .02$ &  \color{black}$.65 \scriptscriptstyle \pm .02$ &               $.54 \scriptscriptstyle \pm .01$ &               $.23 \scriptscriptstyle \pm .01$ &  $.62 \scriptscriptstyle \pm .09$ &  $.24 \scriptscriptstyle \pm .04$ &  $.51 \scriptscriptstyle \pm .05$ &  $.29 \scriptscriptstyle \pm .03$ &               $.62 \scriptscriptstyle \pm .05$ &               $.28 \scriptscriptstyle \pm .03$ &               $.56 \scriptscriptstyle \pm .07$ &               $.23 \scriptscriptstyle \pm .03$ \\
    &     & FMN. &               $.92 \scriptscriptstyle \pm .02$ &               $.95 \scriptscriptstyle \pm .01$ &               $.86 \scriptscriptstyle \pm .00$ &               $.92 \scriptscriptstyle \pm .00$ &  $.82 \scriptscriptstyle \pm .02$ &  $.88 \scriptscriptstyle \pm .03$ &  $.69 \scriptscriptstyle \pm .02$ &  $.77 \scriptscriptstyle \pm .03$ &  \color{black}$.94 \scriptscriptstyle \pm .01$ &  \color{black}$.96 \scriptscriptstyle \pm .01$ &               $.92 \scriptscriptstyle \pm .02$ &               $.95 \scriptscriptstyle \pm .01$ \\
    &     & IDS &  \color{black}$.92 \scriptscriptstyle \pm .01$ &  \color{black}$.89 \scriptscriptstyle \pm .01$ &               $.84 \scriptscriptstyle \pm .01$ &               $.52 \scriptscriptstyle \pm .04$ &  $.45 \scriptscriptstyle \pm .10$ &  $.19 \scriptscriptstyle \pm .06$ &  $.62 \scriptscriptstyle \pm .25$ &  $.42 \scriptscriptstyle \pm .25$ &               $.87 \scriptscriptstyle \pm .01$ &               $.66 \scriptscriptstyle \pm .05$ &               $.86 \scriptscriptstyle \pm .03$ &               $.64 \scriptscriptstyle \pm .13$ \\
    &     & KDD &               $.87 \scriptscriptstyle \pm .04$ &               $.90 \scriptscriptstyle \pm .03$ &  \color{black}$.95 \scriptscriptstyle \pm .00$ &  \color{black}$.95 \scriptscriptstyle \pm .00$ &  $.74 \scriptscriptstyle \pm .04$ &  $.83 \scriptscriptstyle \pm .01$ &  $.86 \scriptscriptstyle \pm .03$ &  $.90 \scriptscriptstyle \pm .01$ &               $.91 \scriptscriptstyle \pm .01$ &               $.93 \scriptscriptstyle \pm .01$ &               $.91 \scriptscriptstyle \pm .03$ &               $.93 \scriptscriptstyle \pm .02$ \\
    &     & MN. &               $.93 \scriptscriptstyle \pm .01$ &  \color{black}$.96 \scriptscriptstyle \pm .01$ &               $.75 \scriptscriptstyle \pm .01$ &               $.79 \scriptscriptstyle \pm .01$ &  $.82 \scriptscriptstyle \pm .04$ &  $.84 \scriptscriptstyle \pm .04$ &  $.69 \scriptscriptstyle \pm .02$ &  $.75 \scriptscriptstyle \pm .01$ &               $.93 \scriptscriptstyle \pm .01$ &               $.95 \scriptscriptstyle \pm .00$ &  \color{black}$.94 \scriptscriptstyle \pm .01$ &  \color{black}$.96 \scriptscriptstyle \pm .01$ \\
    &     & URL &  \color{black}$.94 \scriptscriptstyle \pm .01$ &  \color{black}$.99 \scriptscriptstyle \pm .00$ &               $.92 \scriptscriptstyle \pm .00$ &               $.98 \scriptscriptstyle \pm .00$ &  $.90 \scriptscriptstyle \pm .01$ &  $.97 \scriptscriptstyle \pm .00$ &  $.92 \scriptscriptstyle \pm .01$ &  $.98 \scriptscriptstyle \pm .00$ &               $.92 \scriptscriptstyle \pm .02$ &               $.98 \scriptscriptstyle \pm .00$ &               $.92 \scriptscriptstyle \pm .02$ &               $.98 \scriptscriptstyle \pm .01$ \\
    &     & mean &                              \color{black}$.87$ &                              \color{black}$.79$ &                                           $.80$ &                                           $.66$ &                              $.72$ &                              $.59$ &                              $.69$ &                              $.60$ &                                           $.83$ &                                           $.70$ &                                           $.78$ &                                           $.68$ \\
    &     & p-val &                                                - &                                                - &                                           $.47$ &                                           $.30$ &                              $.05$ &                              $.03$ &                              $.02$ &                              $.03$ &                                           $.47$ &                                           $.30$ &                                           $.30$ &                                           $.30$ \\

\end{tabular}
}

\end{table}

\subsection{Unknown Anomalies}
\label{sec:eval:unknown}
In our final experiment, we evaluated the transfer performance, i.e. how well the knowledge about known anomalies transfers to unknown ones.
In real-world setting, the known anomalies often cover only a small part of all possible anomalies.
Semi-supervised AD methods are expected to use the prior knowledge about known anomalies to detect unknown ones.
We simulate this setting by limiting the training anomaly classes.
Also in this experiment, \GAA{} took the lead with a mean AUC of $87\%$ as shown in \Cref{tab:results:unknown}.
\GAA{} was \percincrease{.83}{.87} better compared to the next best baseline, DevNet.
Except for CoverType and KDD, \GAA{} was better than the unsupervised methods.
This experiment suggested that the raw gradient contains features that generalise across anomaly types.

Throughout our evaluation, we have seen \GAA{} to deliver superior anomaly detection performance under common limitations in AD.
\GAA{} reliably detected known anomalies, yet generalised to unknown ones.
Moreover, the detection performance remained at a high level even under polluted data sets and little prior knowledge about potential anomalies.
With \GAA{} we provide a reliable AD method applicable to a variety of important applications of AD.

\begin{table}[tb]
    \caption{
    Mean performance depending on the number of training steps of the target network.
    We evaluated the detection of known anomalies given 100 anomalous training samples and show the results after five detection runs.
    }
    \label{tab:results:ablation}

\centering
\begin{tabular}{H H H >{\color{gray}}c >{\color{gray}}c | >{\color{gray}}c >{\color{gray}}c | >{\color{gray}}c >{\color{gray}}c | >{\color{gray}}c >{\color{gray}}c }
    &     &   & \multicolumn{2}{c}{1} & \multicolumn{2}{c}{2} & \multicolumn{2}{c}{3} & \multicolumn{2}{c}{4} \\
    &     &  &     AUC &      AP &     AUC &      AP &                  AUC &                   AP &     AUC &      AP \\
\midrule
& & &  $.87$ &  $.77$ &  \color{black}$.93$ &  $.83$ &  \color{black}$.93$ &  \color{black}$.84$ &  $.92$ &  \color{black}$.84$ \\
\end{tabular}

\end{table}
\subsection{Ablation Study}
\label{sec:eval:abl}
In our ablation study, we took a critical look on the temporal component of \GAA{}.
We analysed the gradient of multiple training states of the target AE for anomalous patterns.
Would it have been sufficient to consider a single time step only?
In \Cref{tab:results:ablation}, we evaluated the gradient of 1, 2, 3 and 4 time steps.
For the single time step case, we replaced the LSTM elements by dense layers.
The mean performance considerably decreased when only evaluating a single time step.
In comparison to three time steps, the AUC dropped by \percincrease{.93}{.87}.
A single extra time step improved the performance.
Expanding the number of time steps did not result in further improvements.
We conclude that the temporal gradient distribution contains features important to AD, which are not present in a static one-step analysis.

\section*{Discussion and Future Work}
In \GAA{}, we expanded the analysis of the gradient space of NNs -- to the best of our knowledge -- the first time to semi-supervised AD.
Based on our evaluation, we have seen that the temporal gradient distribution allows to reliably detect anomalous inputs under diverse extents of prior knowledge on several important fields of application.
Due to the end-to-end nature of \GAA{}, it readily integrates in other application areas.
We hope to spark interest in porting our framework to sequential inputs like sensor measurements or video streams.
Moreover, as we have seen related work in other detection areas, e.g. OOD or adversarial detection, we see potential to apply a gradient-based analysis to other important data mining and IT~security applications e.g. deepfake detection.

\section*{Summary}
In this paper, we introduced \GAA{}: a semi-supervised AD method based on the analysis of the temporal gradient distribution of NNs.
\GAA{} showed superior performance in a purely data-driven way, generalising to several important applications of AD.
Our evaluation motivated that \GAA{} is less susceptible to noisy training data than other state-of-the-art AD methods and requires less known anomalies for reliable detection performance.
With \GAA{}, we extend the analysis of the NN's gradient the first time to semi-supervised AD, providing a reliable AD method to researchers and practitioners.

\section*{Ethical Implications}
Data-driven AD reveals data points that differ from the training data distribution.
Underrepresented groups in the training data may cause a bias in the detection results.
In example of the census data set, which we analysed during our evaluation, e.g. the origin of the citizens could be used for the anomaly decision leading to ethical implications.
To this end, we encourage users of \GAA{} and AD in general to thoroughly evaluate potential biases in the data.

\bibliographystyle{abbrv}
\bibliography{references}

\begin{thebibliography}{10}

\bibitem{akcay_ganomaly_2019}
S.~Akcay, A.~Atapour-Abarghouei, and T.~P. Breckon.
\newblock {GANomaly}: {Semi}-supervised {Anomaly} {Detection} via {Adversarial}
  {Training}.
\newblock In C.~V. Jawahar, H.~Li, G.~Mori, and K.~Schindler, editors, {\em
  Computer {Vision} – {ACCV} 2018}, Lecture {Notes} in {Computer} {Science},
  pages 622--637, Cham, 2019. Springer International Publishing.

\bibitem{beggel_robust_2020}
L.~Beggel, M.~Pfeiffer, and B.~Bischl.
\newblock Robust {Anomaly} {Detection} in {Images} {Using} {Adversarial}
  {Autoencoders}.
\newblock In U.~Brefeld, E.~Fromont, A.~Hotho, A.~Knobbe, M.~Maathuis, and
  C.~Robardet, editors, {\em Machine {Learning} and {Knowledge} {Discovery} in
  {Databases}}, Lecture {Notes} in {Computer} {Science}, pages 206--222, Cham,
  2020. Springer International Publishing.

\bibitem{bergman_classification-based_2020}
L.~Bergman and Y.~Hoshen.
\newblock Classification-{Based} {Anomaly} {Detection} for {General} {Data}.
\newblock In {\em International {Conference} on {Learning} {Representations}},
  2020.

\bibitem{blackard_comparative_1999}
J.~A. Blackard and D.~J. Dean.
\newblock Comparative accuracies of artificial neural networks and discriminant
  analysis in predicting forest cover types from cartographic variables.
\newblock {\em Computers and Electronics in Agriculture}, 24(3):131--151, Dec.
  1999.

\bibitem{borghesi_anomaly_2019}
A.~Borghesi, A.~Bartolini, M.~Lombardi, M.~Milano, and L.~Benini.
\newblock Anomaly {Detection} {Using} {Autoencoders} in {High} {Performance}
  {Computing} {Systems}.
\newblock {\em Proceedings of the AAAI Conference on Artificial Intelligence},
  33(01):9428--9433, July 2019.
\newblock Number: 01.

\bibitem{chalapathy_deep_2019}
R.~Chalapathy and S.~Chawla.
\newblock Deep {Learning} for {Anomaly} {Detection}: {A} {Survey}.
\newblock {\em arXiv e-prints}, page arXiv:1901.03407, 2019.
\newblock \_eprint: 1901.03407.

\bibitem{dhaliwal_gradient_2018}
J.~Dhaliwal and S.~Shintre.
\newblock Gradient {Similarity}: {An} {Explainable} {Approach} to {Detect}
  {Adversarial} {Attacks} against {Deep} {Learning}.
\newblock {\em arXiv:1806.10707 [cs]}, June 2018.
\newblock arXiv: 1806.10707.

\bibitem{golan_deep_2018}
I.~Golan and R.~El-Yaniv.
\newblock Deep {Anomaly} {Detection} {Using} {Geometric} {Transformations}.
\newblock In S.~Bengio, H.~Wallach, H.~Larochelle, K.~Grauman, N.~Cesa-Bianchi,
  and R.~Garnett, editors, {\em Advances in {Neural} {Information} {Processing}
  {Systems}}, volume~31. Curran Associates, Inc., 2018.

\bibitem{goyal_drocc_2020-1}
S.~Goyal, A.~Raghunathan, M.~Jain, H.~V. Simhadri, and P.~Jain.
\newblock {DROCC}: {Deep} {Robust} {One}-{Class} {Classification}.
\newblock In {\em Proceedings of the 37th {International} {Conference} on
  {Machine} {Learning}}, pages 3711--3721. PMLR, Nov. 2020.
\newblock ISSN: 2640-3498.

\bibitem{habibi_lashkari_didarknet_2020}
A.~Habibi~Lashkari, G.~Kaur, and A.~Rahali.
\newblock {DIDarknet}: {A} {Contemporary} {Approach} to {Detect} and
  {Characterize} the {Darknet} {Traffic} using {Deep} {Image} {Learning}.
\newblock In {\em 2020 the 10th {International} {Conference} on {Communication}
  and {Network} {Security}}, {ICCNS} 2020, pages 1--13, New York, NY, USA, Nov.
  2020. Association for Computing Machinery.

\bibitem{hendrycks_baseline_2017}
D.~Hendrycks and K.~Gimpel.
\newblock A {Baseline} for {Detecting} {Misclassified} and
  {Out}-of-{Distribution} {Examples} in {Neural} {Networks}.
\newblock In {\em International {Conference} on {Learning} {Representations}},
  2017.

\bibitem{hendrycks_deep_2019}
D.~Hendrycks, M.~Mazeika, and T.~Dietterich.
\newblock Deep {Anomaly} {Detection} with {Outlier} {Exposure}.
\newblock In {\em International {Conference} on {Learning} {Representations}},
  2019.

\bibitem{hochreiter_long_1997}
S.~Hochreiter and J.~Schmidhuber.
\newblock Long {Short}-{Term} {Memory}.
\newblock {\em Neural Computation}, 9(8):1735--1780, Nov. 1997.
\newblock Conference Name: Neural Computation.

\bibitem{huang_esad_2021}
C.~Huang, F.~Ye, P.~Zhao, Y.~Zhang, Y.-F. Wang, and Q.~Tian.
\newblock {ESAD}: {End}-to-end {Deep} {Semi}-supervised {Anomaly} {Detection}.
\newblock In {\em The 32nd {British} {Machine} {Vision} {Conference}}, Oct.
  2021.

\bibitem{huang_importance_2021}
R.~Huang, A.~Geng, and Y.~Li.
\newblock On the {Importance} of {Gradients} for {Detecting} {Distributional}
  {Shifts} in the {Wild}.
\newblock {\em arXiv:2110.00218 [cs]}, Oct. 2021.
\newblock arXiv: 2110.00218.

\bibitem{ioffe_batch_2015}
S.~Ioffe and C.~Szegedy.
\newblock Batch {Normalization}: {Accelerating} {Deep} {Network} {Training} by
  {Reducing} {Internal} {Covariate} {Shift}.
\newblock In {\em Proceedings of the 32nd {International} {Conference} on
  {Machine} {Learning}}, pages 448--456. PMLR, June 2015.
\newblock ISSN: 1938-7228.

\bibitem{kwon_backpropagated_2020}
G.~Kwon, M.~Prabhushankar, D.~Temel, and G.~AlRegib.
\newblock Backpropagated {Gradient} {Representations} for {Anomaly}
  {Detection}.
\newblock In A.~Vedaldi, H.~Bischof, T.~Brox, and J.-M. Frahm, editors, {\em
  Computer {Vision} – {ECCV} 2020}, Lecture {Notes} in {Computer} {Science},
  pages 206--226, Cham, 2020. Springer International Publishing.

\bibitem{kwon_novelty_2020}
G.~Kwon, M.~Prabhushankar, D.~Temel, and G.~AlRegib.
\newblock Novelty {Detection} {Through} {Model}-{Based} {Characterization} of
  {Neural} {Networks}.
\newblock In {\em 2020 {IEEE} {International} {Conference} on {Image}
  {Processing} ({ICIP})}, pages 3179--3183, Oct. 2020.
\newblock ISSN: 2381-8549.

\bibitem{lecun_gradient-based_1998}
Y.~Lecun, L.~Bottou, Y.~Bengio, and P.~Haffner.
\newblock Gradient-based learning applied to document recognition.
\newblock {\em Proceedings of the IEEE}, 86(11):2278--2324, 1998.

\bibitem{lee_open-set_2021}
J.~Lee and G.~AlRegib.
\newblock Open-{Set} {Recognition} {With} {Gradient}-{Based} {Representations}.
\newblock In {\em 2021 {IEEE} {International} {Conference} on {Image}
  {Processing} ({ICIP})}, pages 469--473, Sept. 2021.
\newblock ISSN: 2381-8549.

\bibitem{li_mad-gan_2019}
D.~Li, D.~Chen, B.~Jin, L.~Shi, J.~Goh, and S.-K. Ng.
\newblock {MAD}-{GAN}: {Multivariate} {Anomaly} {Detection} for {Time} {Series}
  {Data} with {Generative} {Adversarial} {Networks}.
\newblock In I.~V. Tetko, V.~Kůrková, P.~Karpov, and F.~Theis, editors, {\em
  Artificial {Neural} {Networks} and {Machine} {Learning} – {ICANN} 2019:
  {Text} and {Time} {Series}}, Lecture {Notes} in {Computer} {Science}, pages
  703--716, Cham, 2019. Springer International Publishing.

\bibitem{lust_gran_2020}
J.~Lust and A.~P. Condurache.
\newblock {GraN}: {An} {Eﬃcient} {Gradient}-{Norm} {Based} {Detector} for
  {Adversarial} and {Misclassiﬁed} {Examples}.
\newblock In {\em {ESANN} 2020}, page~6, 2020.

\bibitem{mamun_detecting_2016}
M.~S.~I. Mamun, M.~A. Rathore, A.~H. Lashkari, N.~Stakhanova, and A.~A.
  Ghorbani.
\newblock Detecting {Malicious} {URLs} {Using} {Lexical} {Analysis}.
\newblock In J.~Chen, V.~Piuri, C.~Su, and M.~Yung, editors, {\em Network and
  {System} {Security}}, Lecture {Notes} in {Computer} {Science}, pages
  467--482, Cham, 2016. Springer International Publishing.

\bibitem{montazerishatoori_detection_2020}
M.~MontazeriShatoori, L.~Davidson, G.~Kaur, and A.~H. Lashkari.
\newblock Detection of {DoH} {Tunnels} using {Time}-series {Classification} of
  {Encrypted} {Traffic}.
\newblock In {\em The 5th {IEEE} {Cyber} {Science} and {Technology}
  {Congress}}, pages 63--70, Aug. 2020.

\bibitem{pang_learning_2018-1}
G.~Pang, L.~Cao, L.~Chen, and H.~Liu.
\newblock Learning {Representations} of {Ultrahigh}-dimensional {Data} for
  {Random} {Distance}-based {Outlier} {Detection}.
\newblock In {\em Proceedings of the 24th {ACM} {SIGKDD} {International}
  {Conference} on {Knowledge} {Discovery} \& {Data} {Mining}}, {KDD} '18, pages
  2041--2050, New York, NY, USA, July 2018. Association for Computing
  Machinery.

\bibitem{pang_deep_2021}
G.~Pang, C.~Shen, L.~Cao, and A.~V.~D. Hengel.
\newblock Deep {Learning} for {Anomaly} {Detection}: {A} {Review}.
\newblock {\em ACM Computing Surveys}, 54(2):38:1--38:38, Mar. 2021.

\bibitem{pang_deep_2019}
G.~Pang, C.~Shen, and A.~van~den Hengel.
\newblock Deep {Anomaly} {Detection} with {Deviation} {Networks}.
\newblock In {\em Proceedings of the 25th {ACM} {SIGKDD} {International}
  {Conference} on {Knowledge} {Discovery} \& {Data} {Mining}}, {KDD} '19, pages
  353--362, New York, NY, USA, July 2019. Association for Computing Machinery.

\bibitem{pozzolo_calibrating_2015}
A.~D. Pozzolo, O.~Caelen, R.~A. Johnson, and G.~Bontempi.
\newblock Calibrating {Probability} with {Undersampling} for {Unbalanced}
  {Classification}.
\newblock In {\em 2015 {IEEE} {Symposium} {Series} on {Computational}
  {Intelligence}}, pages 159--166, Dec. 2015.

\bibitem{ruff_unifying_2021}
L.~Ruff, J.~R. Kauffmann, R.~A. Vandermeulen, G.~Montavon, W.~Samek, M.~Kloft,
  T.~G. Dietterich, and K.-R. Müller.
\newblock A {Unifying} {Review} of {Deep} and {Shallow} {Anomaly} {Detection}.
\newblock {\em Proceedings of the IEEE}, pages 1--40, 2021.

\bibitem{ruff_deep_2020}
L.~Ruff, R.~A. Vandermeulen, N.~Görnitz, A.~Binder, E.~Müller, K.-R. Müller,
  and M.~Kloft.
\newblock Deep {Semi}-{Supervised} {Anomaly} {Detection}.
\newblock In {\em International {Conference} on {Learning} {Representations}},
  2020.

\bibitem{salehi_unified_2021}
M.~Salehi, H.~Mirzaei, D.~Hendrycks, Y.~Li, M.~H. Rohban, and M.~Sabokrou.
\newblock A {Unified} {Survey} on {Anomaly}, {Novelty}, {Open}-{Set}, and
  {Out}-of-{Distribution} {Detection}: {Solutions} and {Future} {Challenges}.
\newblock {\em arXiv:2110.14051 [cs]}, Oct. 2021.
\newblock arXiv: 2110.14051.

\bibitem{schlegl_f-anogan_2019}
T.~Schlegl, P.~Seeböck, S.~M. Waldstein, G.~Langs, and U.~Schmidt-Erfurth.
\newblock f-{AnoGAN}: {Fast} unsupervised anomaly detection with generative
  adversarial networks.
\newblock {\em Medical Image Analysis}, 54:30--44, May 2019.

\bibitem{schulze_da3g_2021}
J.-P. Schulze, P.~Sperl, and K.~Böttinger.
\newblock {DA3G}: {Detecting} {Adversarial} {Attacks} by {Analysing}
  {Gradients}.
\newblock In E.~Bertino, H.~Shulman, and M.~Waidner, editors, {\em Computer
  {Security} – {ESORICS} 2021}, Lecture {Notes} in {Computer} {Science},
  pages 563--583, Cham, 2021. Springer International Publishing.

\bibitem{sharafaldin_toward_2018}
I.~Sharafaldin, A.~H. Lashkari, and A.~A. Ghorbani.
\newblock Toward generating a new intrusion detection dataset and intrusion
  traffic characterization.
\newblock In {\em {ICISSP}}, pages 108--116, 2018.

\bibitem{sohn_learning_2020}
K.~Sohn, C.-L. Li, J.~Yoon, M.~Jin, and T.~Pfister.
\newblock Learning and {Evaluating} {Representations} for {Deep} {One}-{Class}
  {Classification}.
\newblock Sept. 2020.

\bibitem{sperl_activation_2021}
P.~Sperl, J.-P. Schulze, and K.~Böttinger.
\newblock Activation {Anomaly} {Analysis}.
\newblock In F.~Hutter, K.~Kersting, J.~Lijffijt, and I.~Valera, editors, {\em
  Machine {Learning} and {Knowledge} {Discovery} in {Databases}}, Lecture
  {Notes} in {Computer} {Science}, pages 69--84. Springer International
  Publishing, 2021.

\bibitem{sun_gradient-based_2021}
J.~Sun, L.~Yang, J.~Zhang, F.~Liu, M.~Halappanavar, D.~Fan, and Y.~Cao.
\newblock Gradient-based {Novelty} {Detection} {Boosted} by {Self}-supervised
  {Binary} {Classification}.
\newblock {\em arXiv:2112.09815 [cs]}, Dec. 2021.
\newblock arXiv: 2112.09815.

\bibitem{tavallaee_detailed_2009}
M.~Tavallaee, E.~Bagheri, W.~Lu, and A.~A. Ghorbani.
\newblock A detailed analysis of the {KDD} {CUP} 99 data set.
\newblock In {\em 2009 {IEEE} {Symposium} on {Computational} {Intelligence} for
  {Security} and {Defense} {Applications}}, pages 1--6, July 2009.
\newblock ISSN: 2329-6275.

\bibitem{vu_anomaly_2019}
H.~S. Vu, D.~Ueta, K.~Hashimoto, K.~Maeno, S.~Pranata, and S.~M. Shen.
\newblock Anomaly {Detection} with {Adversarial} {Dual} {Autoencoders}.
\newblock {\em arXiv:1902.06924 [cs]}, Feb. 2019.
\newblock arXiv: 1902.06924.

\bibitem{wilcoxon_individual_1992}
F.~Wilcoxon.
\newblock Individual {Comparisons} by {Ranking} {Methods}.
\newblock In S.~Kotz and N.~L. Johnson, editors, {\em Breakthroughs in
  {Statistics}: {Methodology} and {Distribution}}, Springer {Series} in
  {Statistics}, pages 196--202. Springer, New York, NY, 1992.

\bibitem{woods_comparative_1993}
K.~S. Woods, C.~C. Doss, K.~W. Bowyer, J.~L. Solka, C.~E. Priebe, and W.~P.
  Kegelmeyer.
\newblock Comparative evaluation of pattern recognition techniques for
  detection of microcalcifications in mammography.
\newblock {\em International Journal of Pattern Recognition and Artificial
  Intelligence}, 07(06):1417--1436, Dec. 1993.

\bibitem{xiao_fashion-mnist_2017}
H.~Xiao, K.~Rasul, and R.~Vollgraf.
\newblock Fashion-{MNIST}: a {Novel} {Image} {Dataset} for {Benchmarking}
  {Machine} {Learning} {Algorithms}.
\newblock {\em arXiv:1708.07747 [cs, stat]}, Sept. 2017.
\newblock arXiv: 1708.07747.

\bibitem{yamanaka_autoencoding_2019}
Y.~Yamanaka, T.~Iwata, H.~Takahashi, M.~Yamada, and S.~Kanai.
\newblock Autoencoding {Binary} {Classifiers} for {Supervised} {Anomaly}
  {Detection}.
\newblock In A.~C. Nayak and A.~Sharma, editors, {\em {PRICAI} 2019: {Trends}
  in {Artificial} {Intelligence}}, Lecture {Notes} in {Computer} {Science},
  pages 647--659, Cham, 2019. Springer International Publishing.

\bibitem{ye_understanding_2021}
Z.~Ye, Y.~Chen, and H.~Zheng.
\newblock Understanding the {Effect} of {Bias} in {Deep} {Anomaly} {Detection}.
\newblock volume~3, pages 3314--3320, Aug. 2021.
\newblock ISSN: 1045-0823.

\end{thebibliography}
\end{document}